\documentclass[acmsmall,nonacm]{acmart}
\acmConference[ProLaLa 2023]{POPL 2023 - ProLaLa 2023 - Workshop on Programming Languages and the Law}{December 15, 2023}{Boston, MA, USA}
\begin{document}
\title{Towards an Automatic Consolidation of French Law}
\author{Georges-André Silber}
\email{georges-andre.silber@minesparis.psl.eu}
\affiliation{%
  \institution{CRC, Mines Paris, PSL University}
  \streetaddress{1 rue Claude Daunesse}
  \city{Sophia Antipolis}
  \country{France}
  \postcode{06904}
}

\maketitle

\section{Introduction}
\label{sec:intro}

The texts that are part of French law are modified by amending texts
published in the Official Journal of the French Republic (JORF). The
life cycle of a legislative or regulatory text begins with the
publication of its complete version in the JORF and continues with the
possible publication of texts amending it. The full amended text,
called its \textit{consolidated} version, is never published in the
JORF and has no legal value: only the initial version and the suite of
the ordered modifications of the text are authentic~\cite{SGG2017}.

Since 2008, the French Légifrance~\cite{LEGIFRANCE} website presents
most of the legal texts in their original versions as well as in their
successive versions, consequences of the modifications brought to
these texts over time. The operator of the Légifrance website, the
Direction of Legal and Administrative Information (DILA), manually
reports the modifications described in natural language in the texts
in order to obtain, at each modification date, the complete
consolidated version of the text.

This convenience of access to the texts in an easier-to-read and
easier-to-use version has de facto changed the status of these
consolidated versions: they are seen by most users, including legal
professionals, as the reflection of the applicable
law~\cite{GirardotTX2014}. Moreover, the drafters of new texts, in the
French parliament or in the ministries, start from this consolidated
version to conceive the modifying texts. It is therefore of the utmost
importance that this consolidation work be free of errors and
available as soon as possible.

We present here preliminary work integrated in the Legistix tool that
we are developing, whose objective is to create an automated and
reliable consolidation system for French legal texts. This work is
based both on regular expressions used in several compound grammars,
similar to the successive passes of a compiler, and on a new
specialized language of functional type, allowing to describe the
changes applied to the texts.

For each modifier text, our tool generates fully automatically a
computer program in this new language which, when executed, performs
the changes induced by the modifier text on the target texts. In
previous work on this topic, presented for example
in~\cite{FabriziS2021}, only the problem of classifying types of
modification is addressed. To the best of our knowledge, our work is
the first to present a complete approach to identify target texts and
to transform the natural language instructions of the modifier text
into a computer program formalizing the actual transformation rules.

Unlike other topics of study related to legal texts, such as
extracting automatically reasonings, rules of law or trying to
understand the semantics of the text, we place ourselves in a purely
\textit{legistic}~\cite{SGG2017} perspective, i.e. we do not try to
understand the meaning of the text, but only its structure and its
relations to other texts. Since our problem is clearly defined, it
lends itself well to automation, since the consolidation must be
careful not to interpret the changes but only to apply them to the
letter~\cite{GirardotTX2014}.

\section{Example of consolidation}
\label{sec:exemple}

In order to illustrate the complexity of a tool like Legistix, we will
use in this document an example of consolidation based on the law
2022-1348 of 24 October 2022 published in the JORF of 25 October 2022
and presented fig.~\ref{fig:loi-2022-1348-art-1}. Like all texts
published in the JORF, unless otherwise indicated, this law came into
force the day after its publication, i.e. on October 26, 2022. We can
also note in paragraph III of the law a \textit{delayed entry into
  force} of some induced changes on January 1, 2023. The text targeted
by the changes is article L723-4 of the Commercial Code is presented
in fig.~\ref{fig:code:commerce:L723-4:20211013}.\footnote{These
  changes, applied manually by the DILA, can be viewed online on the
  Légifrance website:
  \url{https://www.legifrance.gouv.fr/codes/section_lc/LEGITEXT000005634379/LEGISCTA000006161381/2022-10-26/\#LEGISCTA000006161381}}.

\begin{figure}[ht]
  \hrule
  \begin{center}
    \textbf{Article 1 of law 2022-1348 of october 24, 2022 (partial)}
  \end{center}
  {\small\begin{itemize}
    \item[I.] - Article L723-4 of the Commercial Code is amended as
      follows:
      \begin{itemize}
      \item[1°] At the beginning of the first paragraph, the following
        is added: "\textit{I. -}";
      \item[2°] In 1°, the second occurrence of the word:
        "\textit{and}" is replaced by the word: "\textit{or}";
      \item[3°] In 3° and 4°, after the word: "\textit{judicial}",
        if inserted the word: "\textit{rescue,}";
      \item[4°] In 4° bis, the first occurrence of the word:
        "\textit{were}" is suppressed;
      \item[5°] In 5°, after the word: "\textit{qualities}", are
        inserted the words: "\textit{and duties}";
      \item[6°] The last paragraph is replaced by a II redacted as
        follows:
        \begin{flushleft}
          "\textit{II. - Also eligible are, [...]}\\
          "\textit{2° [...] establishments registered in the trade
            directory or within the jurisdiction of the courts. [...]
          }"
        \end{flushleft}
      \end{itemize}
    \item[II.] - In the first sentence of 2° of II of article L723-4
      of the French Commercial Code, as amended by I of this article,
      the words: "\textit{trade directory}" are replaced by the words:
      "\textit{national register of company or establishment in the
        trades and crafts sector}".
    \item[III.] - II of this article shall apply as of January 1,
      2023. This act shall be executed as a law of the of the state.
    \end{itemize}}
    \hrule
    \caption{Partial reproduction of the unique article of the law
      2022-1348 of october 24, 2022 published in the official journal
      of october 25, 2022~\cite{JORFARTI000046480974}, with the
      identifier \texttt{fr/loi/2022-1348/1/20221025}. Non official
      translation from the French. Missing text is between [...].}
    \label{fig:loi-2022-1348-art-1}
\end{figure}

\begin{figure}[ht]
  \hrule
  \begin{center}
    \textbf{Article L723-4 of the French Commercial Code, in force
      since october 13, 2021}
  \end{center}
  {\small\begin{flushleft}
      Persons who are at least thirty years of age shall
      be eligible for election as a judge of a commercial court:\\
    \begin{itemize}
    \item[1°] Registered on the electoral lists of the chambers of
      commerce and industry and the chambers of trade and crafts [...]
            
    \item[3°] In respect of which a judicial recovery or liquidation
      procedure [...]
      
    \item[4°] [...] a judicial recovery or liquidation procedure is in
      progress on the day of the vote;
      
    \item[4° bis] Who were not were subject to the sanctions [...]
            
    \item[5°] And that justify [...] of the qualities listed [...]
    \end{itemize}
    
    The following are also eligible for election as members of the
    commercial courts, [...].
  \end{flushleft}}
  \hrule
  \caption{Partial reproduction of the article L723-4 of the French
    Commercial Code~\cite{LEGIARTI000044191424}, in force since
    october 13, 2021, with the identifier
    \texttt{fr/code/commerce/L723-4/20211013}. Non official
    translation from the French. Missing text is between [...]}
  \label{fig:code:commerce:L723-4:20211013}
\end{figure}

The amending text that modifies the target code article contains
instructions for humans to modify the original text and to obtain the
new text. These instructions are insertions, deletions, replacements,
etc. We can also note in the 4° bis paragraph of the article a mistake
made by the legislator of a previous amending law voted in the
Parliament: "\textit{Who were not were...}". The objective of our tool
is to transform these instructions intended for humans into a computer
program allowing to apply them automatically.

\section{Data, identifiers and versions}
\label{sec:data}

The data integrated into Legistix are made up of the whole of the
JORF~\cite{DATAJORF} and LEGI~\cite{DATALEGI} databases provided and
updated daily by the DILA. These databases contain all the texts
published in the JORF since 1990 as well as all the texts that have
been consolidated by the DILA. The data are integrated into the
Legistix database on an ongoing basis, thanks to an integration
process that improves the data, reconstructing in particular the
paragraphs and the versions of the tables of contents not provided in
the original data. For example, the Commercial Code has over 600
versions since 2000.

Each document receives a unique identifier in the form of a URI. The
form of this URI is a continuation of the work carried out with the
DILA in the context of the Légimobile~\cite{LarredeP2011}
project. These identifiers have the particularity of being perennial
and of being able to be systematically derived from the information in
the document itself.  Thus, for example, article 1 of the law
2022-1348 of October 24, 2022 presented in
fig.~\ref{fig:loi-2022-1348-art-1} has the unique URI
\url{fr/loi/2022-1348/1}, representing all the versions of this
article over time.

To specify a particular version, the identifiers also support the many
dates governing the life cycle of a law: date of signature, date of
publication in the JORF, dates of modification leading to versions,
date of of abrogation. Thus, the version published in the JORF on
October 25, 2022 of the previous article has as the URI
\url{fr/loi/2022-1348/1/20221025}. Following the same logic, article
L723-4 of the Commercial Code in its version in force since October
13, 2021 presented fig.~\ref{fig:code:commerce:L723-4:20211013} has
the URI \url{fr/code/commerce/L723-4/20211013}.

\section{Detection of modified texts and creation of new versions}
\label{sec:target}

In the example of law 2022-1348, Legistix detects the reference
"\textit{L723-4 of the Commercial Code}" present twice. Thanks to the
identifier mechanism described above, it will be resolved into the
identifier \url{fr/code/commerce/L723-4}, without the need to query an
external resolution mechanism: the character string of the reference
is enough.

Here we can start describing the program obtained automatically by
Legistix: presented here for more clarity in a functional language
with a Python-like syntax, it will allow us to illustrate some of the
mechanisms we are implementing. First of all, the source text
\texttt{s} can be defined as the article 1 of the law 2022-1348 in its
version as of October 25, 2022 as published in the JORF, with
\texttt{db} an object representing the database of the texts and
\texttt{get\_version} a method returning a reference to an existing
version:

\begin{verbatim}
s = db.get_version("fr/loi/2022-1348/1", Date(2022, 10, 25))
# s is fr/loi/2022-1348/1/20221025
\end{verbatim}

Legistix must now find the different versions of the target text that
will be used in the described changes. The first version we need is
the one from which we start, i.e. the version on which the changes are
to be applied. For the first reference found in paragraph I, the date
is the date of publication in the JORF of the source text,
i.e. October 25, 2022. The reference:

\begin{verbatim}
t = db.get_version("fr/code/commerce/L723-4", Date(2022, 10, 25))
# t is fr/code/commerce/L723-4/20211013 (!)
# that is, the version in force on October 25, 2022
\end{verbatim}

\noindent
is positioned in article L723-4 in its version as of October 13,
2021. Indeed, the version in force of this text on October 25, 2022 is
that of October 13, 2021~\cite{LEGIARTI000044191424}, as modified by
Article 1 of Law 2021-1317 which came into force on the same
day~\cite{JORFARTI000044183798}.

As a text in the JORF comes into force by default on the day after its
publication, i.e. here on 26 October 2022, the new version of article
L723-4 described in I of law 2022-1348 must be created on this date
with the function \texttt{new\_version}, from the version represented
by \texttt{t}:

\begin{verbatim}
v1 = t.new_version(Date(2022, 10, 26))
# v1 is fr/code/commerce/L723-4/20221026, created from target t
\end{verbatim}

In II, changes are described with reference to Article L723-4 of the
Commercial Code as revised by I of this article, i.e. \textit{after}
the changes in I have been applied. Combining this information with
the information in III, which indicates that the changes in II are to
apply on January 1, 2023, a new version must be created on that date:

\begin{verbatim}
v2 = v1.new_version(Date(2023, 1, 1))
# v2 is fr/code/commerce/L723-4/20230101, created from version v1
\end{verbatim}

The names \texttt{s}, \texttt{t}, \texttt{v1} and \texttt{v2}
represent abstract references to documents. As we will see in the next
section, these references can be used to abstractly reference portions
of text. For example, the methods \texttt{par} and \texttt{sen} will
allow to generate functions which will respectively return references
to a paragraph or a sentence when they are evaluated. Thus,
\texttt{s.par("I")} generates a function which, when evaluated, will
return a reference to paragraph I of article 1 of law 2022-1348. These
generators can be combined as in:

\begin{verbatim}
v1.par("II").par("2°").sen(1)
\end{verbatim}

\noindent
which generates a function allowing to obtain the first sentence of
the 2° of the II of the article L723-4 of the commercial code in its
version of october 26, 2022.

\section{Detection of changes in the modifying texts}
\label{sec:source}

After identifying the versions of the target texts mentioned in the
modifying texts, Legistix detects the changes and transforms them into
a sequence of functions to apply them. From the text in
fig.~\ref{fig:loi-2022-1348-art-1}, Legistix generates the complete
program shown in fig.~\ref{fig:legistix-loi-2022-1348-art-1}.

\begin{figure}[ht]
\begin{verbatim}
s = db.get_version("fr/loi/2022-1348/1", Date(2022, 10, 25))
t = db.get_version("fr/code/commerce/L723-4", Date(2022, 10, 25))

v1 = t.new_version(Date(2022, 10, 26))
pI = s.par("I")
v1.schedule_changes(s, [
    v1.prepend(pI.par("1"), v1.par(1), "I. -"),
    v1.replace(pI.par("2"), v1.par("1°"), ("and", 2), "or"),
    v1.insert(pI.par("3"), v1.par("3°"), "judicial", "rescue,"),
    v1.insert(pI.par("3"), v1.par("4°"), "judicial", "rescue,"),
    v1.suppress(pI.par("4"), v1.par("4° bis"), ("were", 1)),
    v1.insert(pI.par("5"), v1.par("5°"), "qualities", "and duties"),
    v1.replace_par(pI.par("6"), v1.lastpar(), "II. - Also eligible[...]")
])
v1 = db.add_version(v1.apply_changes())

v2 = v1.new_version(Date(2023, 1, 1))
pII = s.par("II")
v2.schedule_changes(s, [
    v2.replace(pII, v2.par("II").par("2°").sen(1), "trade directory",
               "national register of company[...]")
])
v2 = db.add_version(v2.apply_changes())
\end{verbatim}
  \caption{Legistix program derived from article 1 of the law
    2022-1348 of October 24, 2022, published in the official journal
    of October 25 October 25, 2022~\cite{JORFARTI000046480974}.}
  \label{fig:legistix-loi-2022-1348-art-1}
\end{figure}

It creates two new versions of article L723-4 of the Commercial Code,
\texttt{v1} effective October 26, 2022, with the changes described in
paragraph I of article 1 of law 2022-1348 (\texttt{s.par("I")}),
applying to the version of the Commercial Code as of October 13, 2021,
and \texttt{v2} effective January 1, 2023 with the changes described
in paragraph II (\texttt{s.par("II")}). Note that the second version
is created from the first. The methods \texttt{prepend},
\texttt{replace}, \texttt{insert}, \texttt{suppress},
\texttt{replace\_par} are function generators which are not applied
directly by the program, as long as the method
\texttt{apply\_changes()} is not called.

Each method used in the program translates an operation from natural
language. Each function describing a change is of the form
\texttt{action(source, target, what...)}, where \texttt{source} and
\texttt{target} indicate respectively the text fragment at the origin
of the change and the target. The parameters \texttt{what} describe
the changes, with for example \texttt{("and", 2)} which represents the
second occurrence of the word "\texttt{and}" and "\texttt{rescue}" a
replacement word.

\section{Conclusion and perspectives}

We started with a reference database of several decades where the
consolidation was already carried out manually. This historical
consolidation serves as a reference to measure the reliability of our
approach. The Legistix tool is able to automate 93\% of the
consolidation operations that were previously performed manually.  Our
efforts will continue in order to reach a rate of 100\%, by analyzing
the undetected cases, some of which are ambiguous even for a human,
thus contradicting the mechanizable aspect desired by the
legislator~\cite{GirardotTX2014}.

The next step will be to extend our work to the EUR-Lex~\cite{EURLEX}
database containing the law of the European Union and where
regulations and directives are published on a publication model
similar to French law.

Third, we will propose an expert system~\cite{DelaetA2022} to assist
legislators for the drafting of modifying texts where the formal rules
of modification (the program) could be generated directly during the
drafting of the text.  This would set up a virtuous circle where the
impacts of the modifications could be immediately visualized, thus
improving and making the production of law more reliable.

When the 100\% rate will be reached, there will be no delay anymore
between the publication of the modification texts and the availability
of the consolidated versions. The consolidated text will only be a
by-product of the application of a series of programs starting from
the original text. These programs could be voted on at the same time
as the amending text, making the adage \textit{code is
  law}~\cite{LessigL2000} legally real.

\section*{Acknowledgements}

Thanks to Salomé Ouaknine, a student at Mines Paris, for her help
during her two-month research internship on the subject of automatic
consolidation, especially on the clearing of the future hybrid
approach mixing regular expressions and machine learning.

\bibliographystyle{ACM-Reference-Format}
\bibliography{prolala2023}


\begin{thebibliography}{13}


\ifx \showCODEN    \undefined \def \showCODEN     #1{\unskip}     \fi
\ifx \showDOI      \undefined \def \showDOI       #1{#1}\fi
\ifx \showISBNx    \undefined \def \showISBNx     #1{\unskip}     \fi
\ifx \showISBNxiii \undefined \def \showISBNxiii  #1{\unskip}     \fi
\ifx \showISSN     \undefined \def \showISSN      #1{\unskip}     \fi
\ifx \showLCCN     \undefined \def \showLCCN      #1{\unskip}     \fi
\ifx \shownote     \undefined \def \shownote      #1{#1}          \fi
\ifx \showarticletitle \undefined \def \showarticletitle #1{#1}   \fi
\ifx \showURL      \undefined \def \showURL       {\relax}        \fi
\providecommand\bibfield[2]{#2}
\providecommand\bibinfo[2]{#2}
\providecommand\natexlab[1]{#1}
\providecommand\showeprint[2][]{arXiv:#2}

\bibitem[DAT(2022a)]%
        {DATAJORF}
\bibfield{editor}{\bibinfo{person}{Direction of Legal} {and}
  \bibinfo{person}{Administrative~Information (DILA)}} (Eds.).
  \bibinfo{year}{2022}\natexlab{a}.
\newblock \bibinfo{booktitle}{\emph{Data in XML format containing all laws and
  regulations published in the Official Journal of the French Republic (JORF)
  since 1990}}.
\newblock
\urldef\tempurl%
\url{https://www.data.gouv.fr/fr/datasets/jorf-les-donnees-de-l-edition-lois-et-decrets-du-journal-officiel/}
\showURL{%
\tempurl}


\bibitem[DAT(2022b)]%
        {DATALEGI}
\bibfield{editor}{\bibinfo{person}{Direction of Legal} {and}
  \bibinfo{person}{Administrative~Information (DILA)}} (Eds.).
  \bibinfo{year}{2022}\natexlab{b}.
\newblock \bibinfo{booktitle}{\emph{Data in XML format containing French law
  consolidated by the Direction of Legal and Administrative Information
  (DILA)}}.
\newblock
\urldef\tempurl%
\url{https://www.data.gouv.fr/fr/datasets/legi-codes-lois-et-reglements-consolides/}
\showURL{%
\tempurl}


\bibitem[EUR(2022)]%
        {EURLEX}
\bibfield{editor}{\bibinfo{person}{Publications~Office of~the European~Union}}
  (Ed.). \bibinfo{year}{2022}\natexlab{}.
\newblock \bibinfo{booktitle}{\emph{EUR-Lex, access to European Union Law}}.
\newblock
\urldef\tempurl%
\url{https://eur-lex.europa.eu}
\showURL{%
\tempurl}


\bibitem[LEG(2022)]%
        {LEGIFRANCE}
\bibfield{editor}{\bibinfo{person}{Direction of Legal} {and}
  \bibinfo{person}{Administrative~Information (DILA)}} (Eds.).
  \bibinfo{year}{2022}\natexlab{}.
\newblock \bibinfo{booktitle}{\emph{Légifrance}}.
\newblock
\urldef\tempurl%
\url{https://www.legifrance.gouv.fr}
\showURL{%
\tempurl}


\bibitem[Dela{\"e}t et~al\mbox{.}(2022)]%
        {DelaetA2022}
\bibfield{author}{\bibinfo{person}{Alain Dela{\"e}t}, \bibinfo{person}{Denis
  Merigoux}, {and} \bibinfo{person}{Aymeric Fromherz}.}
  \bibinfo{year}{2022}\natexlab{}.
\newblock \showarticletitle{{Turning Catala into a Proof Platform for the
  Law}}. In \bibinfo{booktitle}{\emph{{POPL 2022 - Programming Languages and
  the Law}}}. \bibinfo{address}{Philadelphia, United States}.
\newblock
\urldef\tempurl%
\url{https://hal.inria.fr/hal-03447072}
\showURL{%
\tempurl}


\bibitem[Fabrizi et~al\mbox{.}(2022)]%
        {FabriziS2021}
\bibfield{author}{\bibinfo{person}{Samuel Fabrizi}, \bibinfo{person}{Maria
  Iacono}, \bibinfo{person}{Tesei Andrea}, {and} \bibinfo{person}{Lorenzo
  De~Mattei}.} \bibinfo{year}{2022}\natexlab{}.
\newblock \showarticletitle{A First Step Towards Automatic Consolidation of
  Legal Acts: Reliable Classification of Textual Modifications}. In
  \bibinfo{booktitle}{\emph{Proceedings of the Eighth Italian Conference on
  Computational Linguistics}}. \bibinfo{address}{Milan, Italy}.
\newblock
\urldef\tempurl%
\url{http://ceur-ws.org/Vol-3033/paper26.pdf}
\showURL{%
\tempurl}


\bibitem[Girardot(2014)]%
        {GirardotTX2014}
\bibfield{author}{\bibinfo{person}{Thierry-Xavier Girardot}.}
  \bibinfo{year}{2014}\natexlab{}.
\newblock \showarticletitle{Accéder au droit: importance et défis de la
  consolidation}.
\newblock \bibinfo{journal}{\emph{Documentaliste -- Sciences de l'Information}}
  \bibinfo{volume}{51}, \bibinfo{number}{4} (\bibinfo{year}{2014}),
  \bibinfo{pages}{30--32}.
\newblock
\urldef\tempurl%
\url{https://www.cairn.info/revue-documentaliste-sciences-de-l-information-2014-4-page-30.htm}
\showURL{%
\tempurl}


\bibitem[général~du gouvernement and d'État(2017)]%
        {SGG2017}
\bibfield{author}{\bibinfo{person}{Secrétariat général~du gouvernement}
  {and} \bibinfo{person}{Conseil d'État}.} \bibinfo{year}{2017}\natexlab{}.
\newblock \bibinfo{booktitle}{\emph{Guide de légistique}}.
\newblock \bibinfo{publisher}{La documentation française}.
\newblock


\bibitem[Larrède and Silber(2011)]%
        {LarredeP2011}
\bibfield{author}{\bibinfo{person}{Pierre Larrède} {and}
  \bibinfo{person}{Georges-André Silber}.} \bibinfo{year}{2011}\natexlab{}.
\newblock \showarticletitle{Un plan de classement des données juridiques
  françaises: l’expérience Légimobile}. In
  \bibinfo{booktitle}{\emph{iExpo2011}}.
\newblock
\urldef\tempurl%
\url{https://www.gfii.fr/uploads/docs/un-plan-de-classement-des-donnees-juridiques-francaises-l-experience-legimobile-laureat-au-prix-de-la-meilleure-contribution-scientifique-i-expo-2011.pdf}
\showURL{%
\tempurl}


\bibitem[Lessig(2000)]%
        {LessigL2000}
\bibfield{author}{\bibinfo{person}{Lawrence Lessig}.}
  \bibinfo{year}{2000}\natexlab{}.
\newblock \showarticletitle{Code Is Law}.
\newblock \bibinfo{journal}{\emph{Harvard Magazine}} (\bibinfo{year}{2000}).
\newblock
\urldef\tempurl%
\url{http://harvardmagazine.com/2000/01/code-is-law.html}
\showURL{%
\tempurl}


\bibitem[Republic(2021a)]%
        {JORFARTI000044183798}
\bibfield{author}{\bibinfo{person}{French Republic}.}
  \bibinfo{year}{2021}\natexlab{a}.
\newblock \bibinfo{booktitle}{\emph{Article 1 of law 2021-1317 of october 11,
  2021, published in the french official journal of october 12, 2021}}.
\newblock
\urldef\tempurl%
\url{https://www.legifrance.gouv.fr/jorf/article_jo/JORFARTI000044183798}
\showURL{%
\tempurl}


\bibitem[Republic(2021b)]%
        {LEGIARTI000044191424}
\bibfield{author}{\bibinfo{person}{French Republic}.}
  \bibinfo{year}{2021}\natexlab{b}.
\newblock \bibinfo{booktitle}{\emph{Article L723-4 of the French Commercial
  Code, in force on October 13, 2021, following the modifications made by art.
  1 of the Law 2021-1317 of October 11, 2021}}.
\newblock
\urldef\tempurl%
\url{https://www.legifrance.gouv.fr/codes/article_lc/LEGIARTI000044191424/2021-10-13/}
\showURL{%
\tempurl}


\bibitem[Republic(2022)]%
        {JORFARTI000046480974}
\bibfield{author}{\bibinfo{person}{French Republic}.}
  \bibinfo{year}{2022}\natexlab{}.
\newblock \bibinfo{booktitle}{\emph{Article 1 of law 2022-1348 of october 24,
  2022, published in the french official journal of october 25, 2022}}.
\newblock
\urldef\tempurl%
\url{https://www.legifrance.gouv.fr/jorf/article_jo/JORFARTI000046480974}
\showURL{%
\tempurl}


\end{thebibliography}
\end{document}